\documentclass[a4paper, 10pt, conference]{ieeeconf}      

\IEEEoverridecommandlockouts                              

\title{\LARGE \bf
Fairness-Aware Federated Learning with Trajectory Shapley Value}

\author{Daniel Kuznetsov and Ziqi Wang%
\thanks{Daniel Kuznetsov is with the Faculty of Mathematics, Ecole Normale Supérieure Paris-Saclay, 4 Av.\ des Sciences, 91190 Gif-sur-Yvette, France.
{\tt\small daniil.kuznetsov@ens-paris-saclay.fr}}%
\thanks{Ziqi Wang is with the Chair for Dynamics, Control, Machine Learning and Numerics -- Alexander von Humboldt Professorship, Department of Mathematics, Friedrich-Alexander-Universit\"at Erlangen-N\"urnberg, Cauerstrasse 11, 91058 Erlangen, Germany. {\tt\small ziqi.wang@fau.de}.}%
\thanks{Published in the 24th European Control Conference (\href{https://ieeexplore.ieee.org/document/11625554}{ECC 2026}).}
}%

\usepackage{color}
\usepackage{cite}
\usepackage{amsmath,amssymb,amsfonts}
\usepackage{mathtools}
\usepackage{graphicx}
\usepackage{textcomp}
\usepackage{xcolor}
\usepackage{algorithm}
\usepackage{algpseudocode}
\makeatletter
\let\NAT@parse\undefined
\renewcommand{\theHALG@line}{\thealgorithm.\arabic{ALG@line}}
\makeatother
\usepackage[colorlinks=true,linkcolor=blue,citecolor=blue,urlcolor=blue]{hyperref}
\usepackage{cleveref}
\usepackage{cuted} 
\usepackage{subcaption}

\newcommand{\dist}{\operatorname{Dist}}

\begin{document}

\bstctlcite{IEEEControlNoURLs}

\makeatletter
\let\titleoriginalmakefntext\@makefntext
\long\def\@makefntext#1{%
  \parindent\z@
  \hangindent 1em
  \hangafter 1
  \noindent\hb@xt@1em{\textbullet\hss}#1}
\maketitle
\let\@makefntext\titleoriginalmakefntext
\makeatother

\begin{abstract}
Federated learning is an emerging distributed paradigm that addresses the challenges posed by heterogeneous, privacy-sensitive data. It enables multiple clients to train a model collaboratively by aggregating their local updates at a server. However, conventional aggregation schemes typically use fixed weights that fail to reflect unequal and time-varying client contributions, leading to biased and unstable learning. To improve fairness and stability, we propose the Trajectory Shapley Value (TSV), a contribution metric that evaluates how each client influences the optimization trajectory of the global model using a validation-based, temporally consistent utility. Building on TSV, we design FedTSV, an adaptive aggregation method that converts per-round evaluations into dynamic client weights, allowing the server to respond to heterogeneous and adversarial participation in real time. Experiments on CIFAR-10 with 20\% and 50\% label-corrupting clients show that FedTSV achieves the highest final test accuracy and the lowest final test loss among the compared methods. Its terminal client weights suppress malicious updates while retaining heterogeneous influence among benign clients, supporting contribution-aware aggregation in the federated settings.
\end{abstract}


\section{Introduction}
Conventional centralized learning has become difficult as data in modern systems are generated across distributed devices and organizations, making central collection costly and subject to privacy, proprietary, and communication constraints. These challenges arise across numerous application domains, including renewable-energy systems, mobile computing, autonomous driving systems, smart healthcare, finance, and the industrial Internet of Things~\cite{li2020federated,kairouz2021advancesa,lin2025hierarchical,li2024federated}. In renewable-energy systems, for example, offshore wind turbines exhibit complex fluid--structure interactions and multiscale operating conditions~\cite{bazilevs2013challenges}. Data-driven methods increasingly complement physics-based models for monitoring, but the data are produced across distributed organizations and remain difficult to aggregate centrally~\cite{otter2022review,liu2025comprehensive}.

To address these challenges, Federated Learning (FL) has been proposed to enable collaborative model training without sharing raw data~\cite{mcmahanCommunicationEfficientLearningDeep2017,karimireddySCAFFOLDStochasticControlled2021,song_fedadmm-insa_2025,wang2023beyond}. In the widely used Federated Averaging (FedAvg) algorithm~\cite{mcmahanCommunicationEfficientLearningDeep2017}, the server samples a subset of clients in each round, and each selected client performs several local stochastic gradient descent (SGD) steps before sending its model update to the server. The server then aggregates the updates with fixed weights, typically uniform or proportional to dataset sizes. Although this strategy is simple and efficient, fixed weights do not reflect time-varying client contributions under data heterogeneity or partial participation. Aggregation can also be distorted by low-quality or malicious clients that exert disproportionate influence on the global model. This creates a need for fair and adaptive weighting that evaluates client contributions in real time and preserves robust cooperation.

Existing client evaluation methods in FL often build on the Shapley value (SV)~\cite{shapley_notes_1951}, a cooperative game-theoretic principle that assigns each participant a fair score reflecting its influence on the collective outcome while accounting for interactions. Exact SV computation scales exponentially in the number of participants and is thus impractical for large networks. Monte Carlo approximations~\cite{wang2019measure} reduce the burden but can lose fairness over time because clients that join late or participate infrequently may be undervalued. This imbalance can destabilize rewards and weaken collaboration~\cite{geimer_volatility_2025}. A gradient-based line of work, represented by the cosine-gradient Shapley value (CGSV)~\cite{xuGradientDrivenRewards2021}, evaluates a client or coalition by the cosine similarity between its update and an average reference update, but this can overemphasize large-magnitude updates and make the score sensitive to scaling. In addition, these approaches do not use validation-based information that may reside on the server. Recent work, such as~\cite{morales2025multi}, exploits server-side data to improve global performance under multi-objective criteria that weight labels differently. These observations motivate a trajectory-based formulation.

To address these issues, we propose the Trajectory Shapley Value (TSV), a contribution metric that evaluates how each client affects the optimization trajectory of the global model instead of focusing only on static loss reduction. At each communication round, the server generates a validation-based reference update by taking several SGD steps on held-out validation data starting from the current global model. For any coalition of participating clients, we compute the average update they produce. A bounded per-round utility increases as the coalition update becomes closer to the validation reference, measured through a normalized Euclidean or angular distance, yielding stable evaluation even near stationary points. We estimate per-round Shapley values from this bounded utility using Monte Carlo sampling and accumulate them across rounds. The cumulative contributions are then converted into nonnegative adaptive aggregation weights for global averaging. This mechanism requires only one client-training pass per round plus a server-side validation pass and captures temporal dependencies along the optimization trajectory under partial participation. Finally, we evaluate FedTSV in the presence of both benign clients and malicious clients performing label corruption attacks. FedTSV achieves the highest final test accuracy and the lowest final test loss among all compared methods, while assigning substantially lower aggregation weights to malicious clients.

The main contributions are summarized as follows:
\begin{itemize}
\item We introduce a novel contribution metric, TSV, which evaluates clients through their influence on the optimization trajectory and provides a temporally consistent alternative to conventional SV-based methods.
\item We develop FedTSV, an adaptive-weighting algorithm that leverages TSV to dynamically adjust aggregation weights and thereby improve fairness and robustness in heterogeneous and adversarial environments.

\item We evaluate FedTSV on CIFAR-10 under statistical heterogeneity, partial participation, and 20\% or 50\% label-corrupting clients, examining both global predictive performance and terminal client weights.
\end{itemize}

The remainder of the paper is organized as follows. Section~\ref{sec:fl_formulation} formulates the FL problem and reviews SV-based contribution measures. Section~\ref{sec:tsv_method} presents TSV and the adaptive-weighting algorithm. Section~\ref{sec:experiments} reports numerical experiments and discusses the results. Section~\ref{sec:conclusion} concludes the paper and outlines future work.

\section{Background and Problem Formulation}
\label{sec:fl_formulation}

In this section, we review the basic notions of FL and its training paradigm. We then state the problem of assessing client contributions during collaborative training and motivate the use of the SV as a principled evaluation tool.

\subsection{Federated Learning}
FL is a paradigm in which learning is performed collaboratively across decentralized devices called \emph{clients} under the coordination of a central \emph{server}. The server orchestrates training without accessing raw data, making FL suitable for privacy-sensitive and communication-constrained settings. Unlike centralized empirical risk minimization, FL must account for distributed data ownership, partial client availability, and communication constraints.

Mathematically, let the training dataset $\mathcal{D}$ be distributed among a set of clients $\mathcal{N} := \{1,2,\dots,n\}$, where client $i \in \mathcal{N}$ holds a local dataset $\mathcal{D}_i$. Each client is associated with a local loss function $F_i(\cdot;\mathcal{D}_i)$ defined over model parameters $\theta \in \mathbb{R}^d$. The global objective in FL is
\begin{equation}
    \min_{\theta \in \mathbb{R}^d}\;
    F(\theta;\mathcal{D})
    \coloneqq
    \sum_{i \in \mathcal{N}} \alpha_i\, F_i(\theta;\mathcal{D}_i),
    \label{eq:fl_objective}
\end{equation}
where $\alpha_i \ge 0$ are the aggregation weights. The most common choice is to set $\alpha_i$ proportional to dataset sizes, \emph{i.e.}, $\alpha_i = |\mathcal{D}_i| \big/ \sum_{j \in \mathcal{N}} |\mathcal{D}_j|$.

Although Problem~\eqref{eq:fl_objective} resembles empirical risk minimization, the crucial distinction is that data remain decentralized and the global model is updated only through server-client communication.

In FL, training proceeds in \emph{communication rounds}. At each round, the server interacts with a subset of clients $\mathcal{S}_t \subseteq \mathcal{N}$. Each participating client performs local training on its private data and returns only model updates to the server, after which the server aggregates these updates to form the next global model.
To be concrete, at each round $t$, each selected client $i \in \mathcal{S}_t$ first initializes its local model $\theta^{t,0}_i = \theta^t$ from the current global model. It then performs $K$ local SGD updates:
\begin{equation}\label{eq:local_update}
\theta^{t,k+1}_i = \theta^{t,k}_i - \eta \, \nabla F_i(\theta^{t,k}_i;\,\xi^{t,k}_i),
\end{equation}
where $\eta$ is the local learning rate and $\xi^{t,k}_i \subseteq \mathcal{D}_i$ is a randomly sampled mini-batch.
After these $K$ steps, each client $i \in \mathcal{S}_t$ sends its final local model $\theta^{t+1}_i \coloneqq \theta^{t,K}_i$ to the server. 

The server then aggregates these models to compute the new global model $\theta^{t+1}$ as
\begin{equation}
\theta^{t+1} = \sum_{i \in \mathcal{S}_t} \frac{\alpha_i}{\sum_{j \in \mathcal{S}_t}\alpha_j} \theta^{t+1}_i.
\label{eq:fedavg_agg}
\end{equation}
The principle of FedAvg is summarized in \Cref{alg:fedavg}.

\begin{algorithm}[t]
\caption{Federated Averaging (FedAvg)~\cite{mcmahanCommunicationEfficientLearningDeep2017}}
\label{alg:fedavg}
\begin{algorithmic}[1]
\State Initialize global model parameters $\theta^0$
\For{each round $t = 0,1, \dots, T-1$}
    \State Server selects a subset of clients $\mathcal{S}_t \subseteq \mathcal{N}$
    \For{each client $i \in \mathcal{S}_t$ \textbf{in parallel}}
        \State Server sends current model $\theta^{t}$ to client $i$
        \State Client initializes local model $\theta^{t,0}_i = \theta^{t}$
        \For{$k = 0$ to $K-1$}
            \State Select a mini-batch $\xi^{t,k}_i$
            \State $\theta^{t,k+1}_i = \theta^{t,k}_i - \eta \, \nabla F_i(\theta^{t,k}_i;\,\xi^{t,k}_i)$
        \EndFor
        \State Client sends $\theta^{t+1}_i \coloneqq \theta^{t,K}_i$ back to the server
    \EndFor
    \State Server aggregates updates:
    \begin{equation*}
        \theta^{t+1} =
        \sum_{i \in \mathcal{S}_t}
        \frac{\alpha_i}{\sum_{j \in \mathcal{S}_t}\alpha_j}\,
        \theta^{t+1}_i
    \end{equation*}
\EndFor
\State \Return $\theta^T$
\end{algorithmic}
\end{algorithm}

\subsection{Contribution Estimation and the Shapley Value}

Assessing each client's contribution is crucial for incentive design and robust aggregation, since informative clients should be rewarded while noisy or adversarial updates should be down-weighted. A suitable metric should therefore account for client interactions while remaining computationally feasible over many rounds.

A classical and principled approach to quantifying such contributions is based on the SV~\cite{shapley_notes_1951}, which measures the average marginal improvement that a participant brings to all possible coalitions. For a finite set of clients $\mathcal{N}$ with $n = |\mathcal{N}|$, the SV of client $i \in \mathcal{N}$ with respect to a utility function $u$ is defined as
\begin{equation}
    \phi_i(u) 
    = 
    \sum_{\mathcal{S} \subseteq \mathcal{N} \setminus \{i\}}
    \frac{|\mathcal{S}|! \, (n - |\mathcal{S}| - 1)!}{n!}
    \big( u(\mathcal{S} \cup \{i\}) - u(\mathcal{S}) \big),
    \label{eq:sv_def}
\end{equation}
where $\phi_i(u)$ is the SV of client $i$ and $u(\mathcal{S})$ is the utility of the global model trained using coalition $\mathcal{S}$. For example, possible choices of the utility function include
\begin{equation}
    u(\mathcal{S}) = \operatorname{acc}\big(\theta_{\mathcal{S}}\big)
    \quad\text{and}\quad
    u(\mathcal{S}) = -F_0\big(\theta_{\mathcal{S}}\big).
\end{equation}
Here, $\operatorname{acc}(\cdot)$ denotes the model accuracy evaluated on a fixed validation dataset $\mathcal{D}_0$, $\theta_{\mathcal{S}}$ is the global model obtained by aggregating the updates from clients in $\mathcal{S}$, and $F_0$ is the loss function defined over the server-side validation dataset $\mathcal{D}_0$. Using such a validation dataset is standard in SV-based FL methods~\cite{liu2022gtg,sun_shapleyfl_2023}.

Exact SV computation is intractable in large-scale FL because it requires evaluating all $2^n$ coalitions or averaging over all client permutations. Two complementary strategies alleviate this difficulty: the first is structural decomposition of the utility and the other is sampling-based approximation of marginal contributions.

The first strategy leverages the iterative nature of FL by decomposing the overall utility into per-round contributions:
\begin{equation}
    u(\mathcal{S}) = \sum_{t=0}^{T-1} u^t(\mathcal{S}),
\end{equation}
where each $u^t$ quantifies incremental improvement in validation performance during round $t$. This additive formulation assumes that per-round gains are approximately independent and enables online estimation. By linearity of the SV~\cite{peters_game_2008}, the total contribution is
\begin{equation}
    \phi_i(u) = \sum_{t=0}^{T-1}\phi_i(u^t), \qquad i \in \mathcal{N},
    \label{eq:decomposition}
\end{equation}
which substantially reduces the computational burden and aligns naturally with the per-round structure of FL\@.

Complementing this structural simplification, Monte Carlo sampling techniques have been developed to approximate the marginal contributions in~\eqref{eq:sv_def}. These methods reduce the number of model evaluations from the exponential $\mathcal{O}(2^n)$ complexity of exact computation to $\mathcal{O}(n \log^2 n)$ while maintaining a controllable approximation error~\cite{zhang_efficient_2023,jia2019towards}. Despite these advances, most alternative approaches still lack the fairness guarantees inherent in the SV and often fail to capture synergistic or redundant relationships among heterogeneous clients' data.

\section{Federated Learning with the Trajectory Shapley Value}
\label{sec:tsv_method}
In this section, we introduce TSV, a contribution metric tailored to iterative FL optimization. TSV evaluates clients by how consistently their coalition updates follow a validation-induced descent trajectory, which is especially relevant under non-IID or adversarial participation. We then transform per-round TSV estimates into adaptive aggregation weights, resulting in the FedTSV algorithm.
  
\subsection{Trajectory Shapley Value}

The TSV proposed in this work provides a geometric and optimization-based interpretation of client contributions in FL\@. Since FL is round-dependent, the same update can have different effects across rounds. Comparing coalition updates with a server-side validation reference connects cooperative-game evaluation with the geometry of descent directions.
Because the learning process can be viewed as an optimization problem solved through gradient descent, it can be regarded as a discrete approximation of the continuous gradient flow:
\begin{equation}
    \theta'(t) = -\nabla F\big(\theta(t)\big),
\end{equation}
where $\theta(t)$ represents the global model parameters at time $t$. 
From this viewpoint, each global round in FL can be interpreted as one discrete step along this descent trajectory.
Accordingly, a coalition is considered more informative if its aggregated update moves the global model in a manner consistent with the descent direction indicated by the server-side validation reference.

At communication round $t$, let $\mathcal{S}_t \subseteq \mathcal{N}$ be the participating clients and let $\mathcal{S} \subseteq \mathcal{S}_t$ be any coalition. On the client side, the aggregated update of coalition $\mathcal{S}$ is
\begin{equation}
    \Delta_{\mathcal{S}}^t \coloneqq
    \begin{cases}
        \frac{1}{|\mathcal{S}|}\sum_{i \in \mathcal{S}} \big( \theta^{t+1}_i - \theta^{t} \big), & \mathcal{S} \neq \emptyset,\\
        0, & \mathcal{S} = \emptyset,
    \end{cases}
    \label{eq:coalition_update}
\end{equation}
where $\theta^{t+1}_i$ denotes the local model obtained by client $i$ after performing $K$ steps of local SGD starting from the global model $\theta^{t}$, as described in~\eqref{eq:local_update}.

On the server side, a reference update is computed on held-out validation data as
\begin{equation}
    \Delta_{\mathrm{val}}^t \coloneqq \theta_0^{t+1} - \theta^{t},
\end{equation}
where $\theta_0^{t+1} \coloneqq \theta_0^{t, K}$ is the model obtained after the same number of SGD steps on validation data, initialized from $\theta^{t}$ and optimized with respect to $F_0$. This reference serves as a baseline for evaluating the quality of client-side updates and their consistency with the validation descent direction.

Based on these quantities, we define a per-round TSV utility $v^t: \mathcal{P}(\mathcal{S}_t) \to (0,1]$, which replaces the accuracy-based utility $u^t$ used in conventional evaluation schemes.
The proposed TSV utility is given by
\begin{equation}
    v^t(\mathcal{S})
    =
    \left(
    1 +
    \frac{\dist\!\big(\Delta_{\mathcal{S}}^t,\,\Delta_{\mathrm{val}}^t\big)^2}{\sigma^t}
    \right)^{-1},
    \label{eq:tsv_utility}
\end{equation}
where $\dist(\cdot,\cdot)$ denotes a distance metric on the model parameter space.
We use the Euclidean metric $\dist(a,b)=\lVert a-b\rVert_2$ in this work, although other choices, such as angular distance, are possible.
The positive scalar $\sigma^t$ serves as a round-dependent normalization factor that ensures dimensionless scaling and mitigates numerical instability.
A practical choice is $\sigma^t = \max\{\lVert \Delta_{\mathrm{val}}^t\rVert_2^2,\varepsilon\}$ with a sufficiently small $\varepsilon>0$, which preserves scale information while avoiding division by zero near stationary points. Accumulating the SV induced by $v^t$ across rounds then yields a cumulative measure of proximity to the validation-based trajectory.

This formulation presents several advantages for FL evaluation.
First, in each round, the utility $v^t(\mathcal{S})$ is bounded in $(0,1]$, which enables efficient Monte Carlo estimation of the SV~\cite{zhang_efficient_2023,jia2019towards}.
Second, $v^t(\mathcal{S})$ increases monotonically as the coalition update $\Delta_{\mathcal{S}}^t$ becomes closer to the reference update $\Delta_{\mathrm{val}}^t$, capturing geometric alignment along the optimization trajectory.
Finally, the normalization factor $\sigma^t$ improves robustness by compensating for scale changes that may occur near stationary points.
If a client $i$ is not included in $\mathcal{S}_t$, its marginal contribution in round $t$ is zero, \emph{i.e.}, $\phi_i(v^t)=0$.

\subsection{FedTSV Algorithm with Adaptive Aggregation Weights}

To enhance FL robustness, FedTSV adapts client aggregation weights according to estimated contributions. The weights, therefore, reflect the empirical usefulness of updates along the training trajectory, rather than being fixed by uniform rules or local dataset sizes. This mechanism assigns higher importance to clients with high-quality data while down-weighting potentially malicious participants, often called Byzantine clients~\cite{bagdasaryanHowBackdoorFederated2019}, that may attempt data-poisoning or label-corruption attacks. Since TSV is computed per round and accumulated over time, the resulting weights capture both update quality and long-term reliability.

At each communication round $t$, we estimate a per-round contribution score for each client and update its aggregation weight accordingly. Thus, the coefficients $(\alpha_i)_{i \in \mathcal{N}}$ in~\eqref{eq:fl_objective} become time-dependent, denoted by $(\alpha_i^t)_{i \in \mathcal{N}}$.

Specifically, let $\phi_i(v^t)$ denote the SV of client $i$ at round $t$ with respect to the utility function $v^t$. We maintain cumulative contribution variables $(\varphi_i^t)_{i \in \mathcal{N}}$ recursively as
\begin{equation}
    \varphi_i^{t+1} = \varphi_i^t + \phi_i(v^t),
    \qquad
    \varphi_i^0 = 0,
\end{equation}
and derive the adaptive aggregation weights by
\begin{equation}
    \alpha_i^{t+1} \coloneqq \max\big\{0,\, \varphi_i^{t+1}\big\},
\end{equation}
where truncation at zero suppresses clients with negative cumulative contributions, preventing them from adversely influencing the global model. If all participating clients have zero adaptive weight in a given round, FedTSV falls back to uniform aggregation for that round.

Because the exact SV $\phi_i(v^t)$ is computationally intractable, we employ a Monte Carlo approximation following the efficient estimation strategy in~\cite{zhang_efficient_2023}. After obtaining the updated weights $\alpha_i^{t+1}$, the server performs weight-based aggregation.
Algorithm~\ref{alg:fedtsv} summarizes the FedTSV procedure.

\begin{algorithm}[t]
\caption{FedTSV: Adaptive Aggregation with TSV}
\label{alg:fedtsv}
\begin{algorithmic}[1]
\State Initialize global model $\theta^0$ and $\varphi_i^0 = 0$ for all $i \in \mathcal{N}$
\For{each round $t = 0, 1, \dots, T-1$}
    \State Server selects a subset of clients $\mathcal{S}_t \subseteq \mathcal{N}$
    \For{each client $i \in \mathcal{S}_t$ \textbf{in parallel}}
        \State Server sends $\theta^{t}$ to client $i$
        \State Client sets $\theta^{t,0}_i = \theta^{t}$
        \For{$k = 0$ to $K-1$}
            \State Choose a random mini-batch $\xi^{t,k}_i \subseteq \mathcal{D}_i$
            \State $\theta^{t,k+1}_i = \theta^{t,k}_i - \eta \, \nabla F_i(\theta^{t,k}_i;\,\xi^{t,k}_i)$
        \EndFor
        \State Client sends $\theta^{t+1}_i \coloneqq \theta^{t,K}_i$ back to the server
    \EndFor
    \State Server computes a validation reference:
    \State Set $\theta_0^{t,0} = \theta^{t}$
    \For{$k = 0$ to $K-1$}
        \State Choose a random mini-batch $\xi^{t,k}_0 \subseteq \mathcal{D}_0$
        \State $\theta_0^{t,k+1} = \theta_0^{t,k} - \eta \, \nabla F_0(\theta_0^{t,k};\,\xi^{t,k}_0)$
    \EndFor
    \State Set $\Delta_{\mathrm{val}}^t \coloneqq \theta_0^{t,K} - \theta^{t}$
    \For{$i \in \mathcal{N}$}
        \If{$i \in \mathcal{S}_t$}
            \State Estimate $\phi_i(v^t)$ using the utility in~\eqref{eq:tsv_utility}
            \State $\varphi_i^{t+1} = \varphi_i^{t} + \phi_i(v^t)$
        \Else
            \State $\varphi_i^{t+1} = \varphi_i^{t}$
        \EndIf
        \State Update $\alpha_i^{t+1} = \max\{0,\varphi_i^{t+1}\}$
    \EndFor
    \State Server aggregates by
  \begin{equation*}
\theta^{t+1} = \sum_{i \in \mathcal{S}_t} \frac{\alpha_i^{t+1}}{\sum_{j\in \mathcal{S}_t}\alpha_j^{t+1}} \theta^{t+1}_i 
\end{equation*}
\EndFor
\State \Return $\theta^T$
\end{algorithmic}
\end{algorithm}

The proposed FedTSV method belongs to a broader class of adaptive-weighting FL algorithms that adjust the weights $(\alpha_i^t)_{i \in \mathcal{N}}$ in~\eqref{eq:fl_objective} over time. The problem is therefore mathematically more delicate because, at each round $t$, the algorithm effectively minimizes a time-varying objective. The convergence analysis for FedTSV is left for future work.

\section{Experimental Results}\label{sec:experiments}

This section evaluates FedTSV against representative FL aggregation and contribution-estimation baselines. The experiments test two complementary properties. First, predictive performance, measured by global accuracy and loss. Second, contribution quality, measured by the learned client weights. Together, these metrics assess whether FedTSV improves optimization while distinguishing reliable IID clients, benign non-IID clients, and malicious clients. The source code is available at \url{https://github.com/DCN-FAU-AvH/FedTSV}.

\subsection{Experimental Setup}\label{subsec:setup}

\noindent\textit{Datasets and client configuration.}
We use CIFAR-10~\cite{krizhevsky2009learning} and a convolutional neural network (CNN) with two convolutional and three fully connected layers. The $n=100$ clients comprise either 50 IID, 30 non-IID, and 20 malicious clients, or 25 IID, 25 non-IID, and 50 malicious clients.
Non-IID label proportions follow a Dirichlet distribution with concentration parameter $0.1$~\cite{hsu_measuring_2019}. Malicious clients apply a fixed cyclic label permutation. The server samples 10 clients uniformly at random per round to form $\mathcal{S}_t$.

\smallskip
\noindent\textit{Training configuration.}
Training runs for $T=500$ communication rounds with batch size $200$ and one local epoch per selected client. For each method, we report the best results from a grid search over local learning rates $\eta\in\{0.005,0.01,0.05,0.1\}$ and server step sizes $\eta_g\in\{0.01,0.1,1\}$. Here, $\eta$ follows~\eqref{eq:local_update}, and $\eta_g$ scales the weighted average of client displacements $\theta_i^{t+1}-\theta^t$ before addition to $\theta^t$. Setting $\eta_g=1$ recovers the model averaging in Algorithms~\ref{alg:fedavg} and~\ref{alg:fedtsv}.
The fixed validation set $\mathcal{D}_0$ contains 500 samples held out from training and disjoint from the test set. Each round uses 20 independently and uniformly sampled permutations of $\mathcal{S}_t$ to estimate $\phi_i(v^t)$.

\smallskip
\noindent\textit{Compared methods.}
The baselines are FedAvg~\cite{mcmahanCommunicationEfficientLearningDeep2017}, CGSV~\cite{xuGradientDrivenRewards2021}, which evaluates alignment with an average update, and leave-one-out (LOO)~\cite{wang2019measure}, which measures the validation-performance change when a client's update is removed. FedTSV instead accumulates Shapley values of the validation-trajectory utility to construct adaptive weights.

\begingroup
\captionsetup{hypcap=false}
\setlength{\textfloatsep}{12pt}
\begin{figure}[t]
    \centering
    \includegraphics[width=0.987\linewidth]{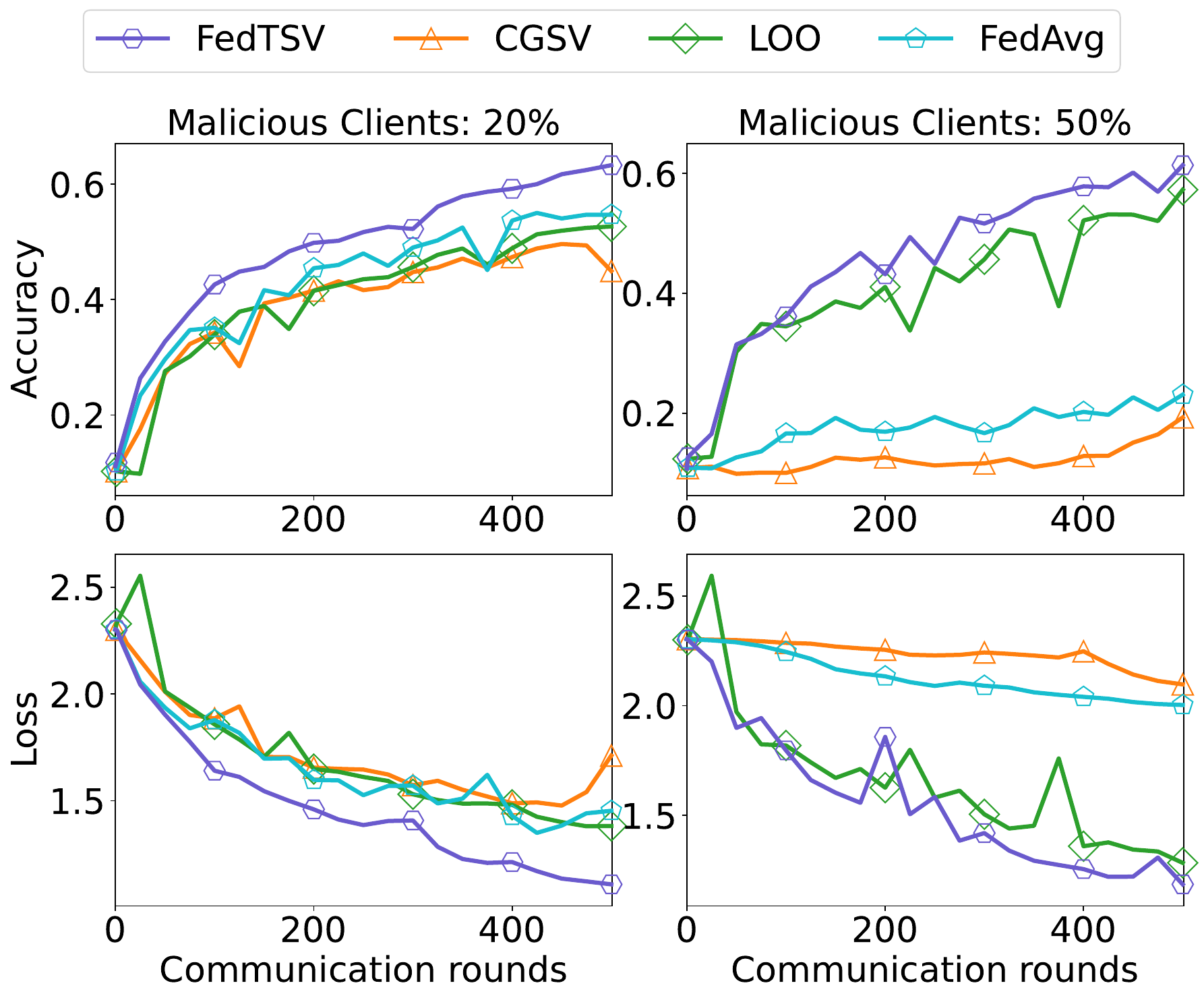}
    \caption{Test accuracy and loss on CIFAR-10 with 20\% (left) and 50\% (right) malicious clients. FedTSV achieves the highest accuracy and the lowest loss in both settings.}
    \label{fig:tsv-client-accuracy}
\end{figure}

\begin{figure}[t]
    \centering
    \begin{subfigure}[t]{\linewidth}
        \centering
        \includegraphics[width=0.987\linewidth]{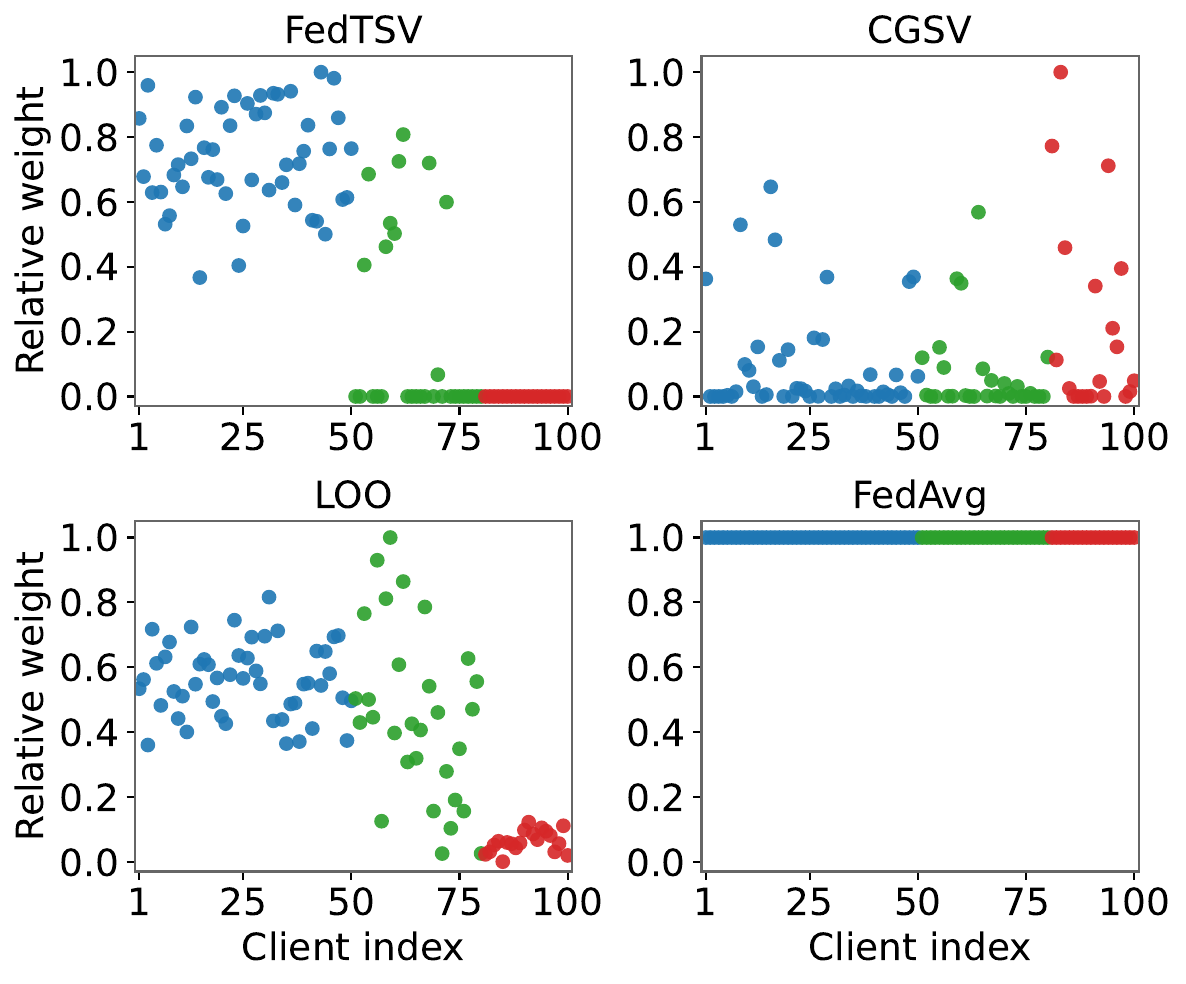}
        \caption{20\% malicious clients.}
        \label{fig:tsv-contrib-20-malicious}
    \end{subfigure}

    \medskip

    \begin{subfigure}[t]{\linewidth}
        \centering
        \includegraphics[width=0.987\linewidth]{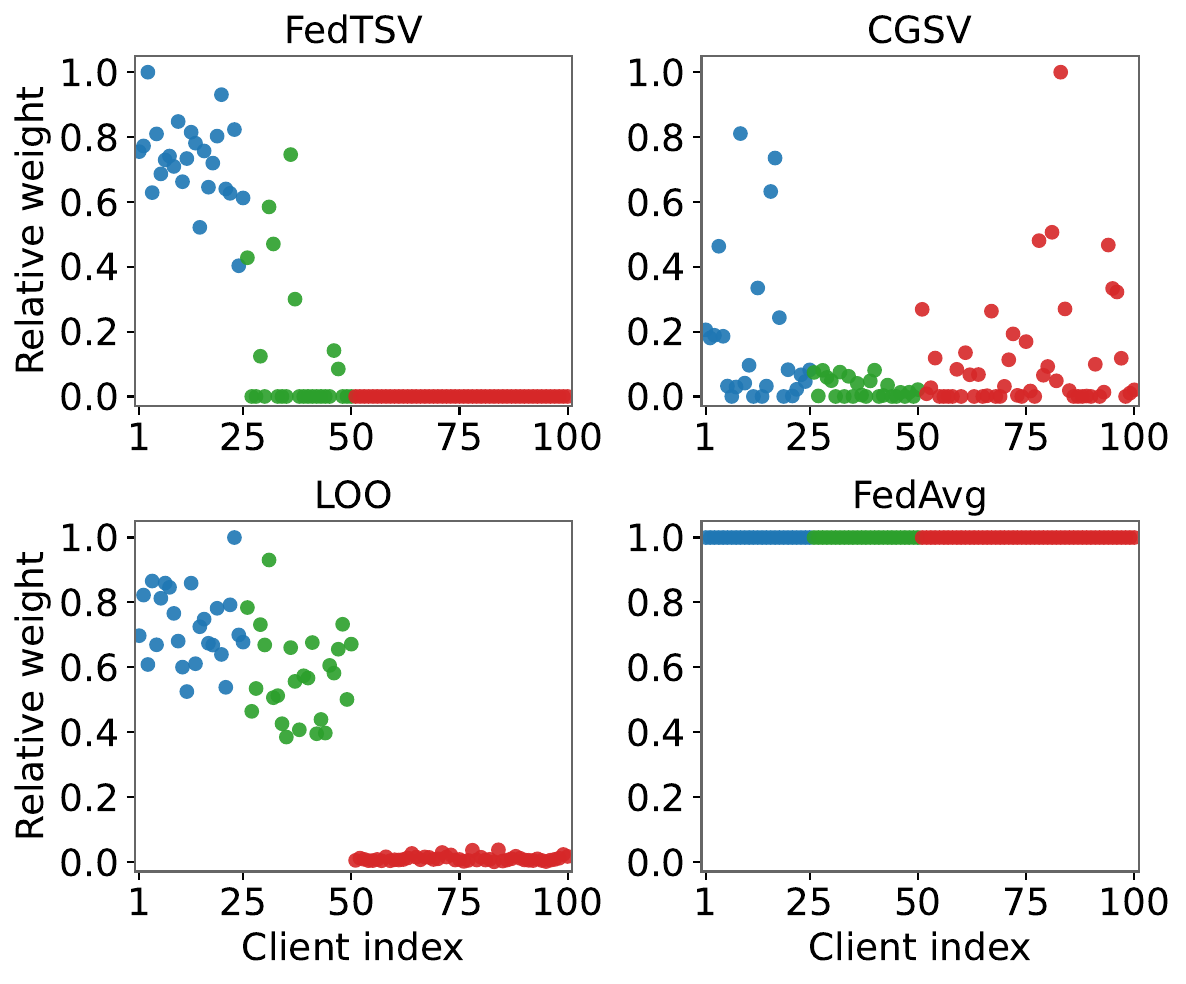}
        \caption{50\% malicious clients.}
        \label{fig:tsv-contrib-50-malicious}
    \end{subfigure}

    \caption{Terminal relative client weights on CIFAR-10. Blue, green, and red points denote IID, non-IID, and malicious clients. The populations comprise (a) 50/30/20 and (b) 25/25/50 clients in these groups, respectively. Each method's terminal weights are normalized by its own maximum.}
    \label{fig:tsv-contrib}
\end{figure}

\subsection{Evaluation of Accuracy and Empirical Robustness}\label{subsec:accuracy}

Figure~\ref{fig:tsv-client-accuracy} reports global test accuracy and loss over 500 communication rounds. With 20\% malicious clients, all four methods initially improve, but their trajectories gradually separate. FedTSV maintains steady progress and reaches the highest final accuracy and lowest loss. FedAvg and LOO also continue to learn but finish at lower accuracy levels. CGSV exhibits greater fluctuation and deteriorates near the end of training, showing weaker stability even when malicious clients form a minority.

With 50\% malicious clients, FedAvg and CGSV make only limited progress after the early rounds. Their accuracy remains comparatively low, while losses stay close to their initial levels. In contrast, LOO and FedTSV continue to improve despite half of the population using corrupted labels. FedTSV finishes with higher accuracy and lower loss than LOO. Its final performance decreases only moderately relative to the 20\% setting, whereas FedAvg and CGSV degrade substantially.

These curves indicate that contribution-aware weighting can mitigate harmful updates in the tested configurations. The distinction is especially clear with 50\% malicious clients, where uniform aggregation struggles to provide a useful training direction.

\subsection{Client Contribution Visualization}\label{subsec:visualization}

Figure~\ref{fig:tsv-contrib} compares terminal relative client weights with 20\% and 50\% malicious clients. Each method's weights are normalized by its own maximum, so the vertical scales indicate relative importance within a method rather than absolute differences across methods.

In both 20\% and 50\% malicious clients cases, the methods differ markedly in how they distinguish client groups. FedAvg assigns identical weights to all clients, preserving the influence of corrupted updates. CGSV assigns near-zero weights to many benign clients while giving several malicious clients substantial weight, including the largest relative weight. This overlap indicates that gradient-direction alignment alone does not reliably distinguish useful from harmful updates in this experiment. LOO achieves stronger separation by assigning appreciable weights to most IID and non-IID clients while keeping malicious weights small and mostly positive.

Notably, FedTSV assigns zero weight to all malicious clients in the 20\% setting, and this suppression persists at 50\%. In both settings, most IID clients retain moderate to high weights, whereas non-IID clients exhibit a wider range. Many non-IID clients receive zero weight, but others retain weights comparable to those of IID clients. This variation is consistent with FedTSV evaluating marginal usefulness relative to the validation reference along the realized trajectory, rather than assigning influence solely by client category. Heterogeneous benign clients can therefore retain substantial influence when their updates are useful, while updates with limited marginal contribution receive little or no weight.

\endgroup

\section{Conclusion}
\label{sec:conclusion}
This work introduced FedTSV, an adaptive-weighting framework for federated learning that employs the Trajectory Shapley Value (TSV) to evaluate client contributions along the optimization trajectory. By integrating trajectory-based update analysis with Shapley value principles, TSV offers a principled mechanism for computing aggregation weights during training. The resulting scheme improves robustness in the presence of heterogeneous and adversarial clients while preserving computational efficiency. Experiments show that FedTSV achieves higher accuracy and faster convergence than existing baselines. Moreover, FedTSV demonstrates superior performance under various levels of client corruption, showcasing its effectiveness under adversarial conditions. A future direction is to investigate the theoretical properties of FedTSV in terms of convergence and fairness guarantees.

\section*{Acknowledgments}
The authors' names are listed in alphabetical order by family name to signify equal contribution. We are grateful to Enrique Zuazua and Roberto Morales for their invaluable advice and discussions. 
This work was funded by the European Union's Horizon Europe MSCA project Modelling and control of flexible structures interacting with fluids (ModConFlex; grant agreement No. 101073558; call HORIZON-MSCA-2021-DN-01-01; project DOI: \href{https://doi.org/10.3030/101073558}{10.3030/101073558}).

\bibliographystyle{IEEEtran}
\bibliography{biblio}

\end{document}